\documentclass{article}
\usepackage[arxiv, final]{neurips_2025}
\usepackage[table]{xcolor}
\usepackage[utf8]{inputenc}
\usepackage[T1]{fontenc}
\usepackage{url}
\usepackage{booktabs}
\usepackage{amsfonts}
\usepackage{textalpha}
\usepackage{nicefrac}
\usepackage{microtype}
\usepackage{xcolor}
\usepackage{graphicx}
\usepackage{caption}
\usepackage{subcaption}
\usepackage{multirow}
\usepackage{makecell}
\usepackage{pifont}
\usepackage{enumitem}
\usepackage{amsmath}
\usepackage{amssymb}
\usepackage{wrapfig}
\usepackage{arydshln}
\usepackage{todonotes}
\setlist[enumerate]{leftmargin=*, label=(\arabic*), topsep=0pt}
\setlist[itemize]{leftmargin=*, topsep=0pt}
\usepackage{tikz}
\definecolor{yellowtext}{RGB}{68,132,243}
\definecolor{yellowred}{RGB}{50,167,82}
\definecolor{yellowblue}{RGB}{251,191,5}

\definecolor{darkblue}{rgb}{0, 0, 0.5}
\definecolor{chocolate}{HTML}{D2691E}
\definecolor{maroon}{HTML}{A00000}
\definecolor{indigo}{HTML}{4B0082}
\definecolor{violet}{HTML}{4B2E83}
\definecolor{lightblue}{rgb}{0.0, 0.0, 0.5}
\definecolor{cadmiumgreen}{rgb}{0.0, 0.42, 0.24}
\definecolor{forestgreen}{rgb}{0.13, 0.55, 0.13}
\usepackage[most]{tcolorbox}
\usepackage{fvextra}
\DefineVerbatimEnvironment{VerbatimWrap}{Verbatim}{
breaklines=true, 
breakindent=0pt, 
breaksymbol={},
fontsize=\scriptsize,
}

\usepackage[framemethod=tikz]{mdframed}
\mdfdefinestyle{customstyle}{
linecolor=cyan!70,
linewidth=3pt,
innerleftmargin=5pt,
topline=false,
rightline=false,
bottomline=false,
leftline=true,
innerrightmargin=5pt,
innertopmargin=5pt,
innerbottommargin=5pt,
backgroundcolor=cyan!8,
}
\newenvironment{custommdframed}
{\begin{mdframed}[style=customstyle]\footnotesize}
{\end{mdframed}}

\usepackage{array}
\newcolumntype{P}[1]{>{\raggedright\arraybackslash}m{#1}}
\usepackage[colorlinks=true,linkcolor=blue,citecolor=green!50!black,urlcolor=magenta!70!black,filecolor=cyan!70!black]{hyperref}

\newcommand{\cmark}{\textcolor{green!50!black}{\ding{51}}}
\newcommand{\xmark}{\textcolor{red!70!black}{\ding{55}}}
\usepackage{pdflscape}
\usepackage{rotating}

\usepackage[table]{xcolor}
\definecolor{lightgray}{rgb}{0.9, 0.9, 0.9}

\usepackage{dcolumn}
\newcolumntype{C}{D{,}{,}{-1}} % center cell contents on commas
\newcolumntype{d}[1]{D{,}{,}{#1}}
\newcolumntype{.}{D{.}{.}{-1}}
\newcolumntype{,}{D{|}{|}{-1}}
\usepackage{lineno}
\title{DiscoVerse: Multi-Agent Pharmaceutical Co-Scientist for Traceable Drug Discovery and Reverse Translation}

\author{%
  Xiaochen Zheng~\thanks{Corresponding author. \url{xiaochen.zheng@roche.com}} \\
  Predictive Modelling\\
  F. Hoffmann-La Roche Ltd.\\
  Basel, Switzerland \\
  \And
  Alvaro Serra \\
  Predictive Modelling\\
  F. Hoffmann-La Roche Ltd.\\
  Basel, Switzerland \\
  \And
  Ilya Schneider Chernov \\
  Predictive Modelling\\
  F. Hoffmann-La Roche Ltd.\\
  Basel, Switzerland \\
  \AND
  Maddalena Marchesi \\
  Clinical Safety\\
  F. Hoffmann-La Roche Ltd.\\
  Basel, Switzerland \\
  \And
  Eunice Musvasva \\
  Translational Safety\\
  F. Hoffmann-La Roche Ltd.\\
  Basel, Switzerland \\
  \And
  Tatyana Y. Doktorova \\
  Predictive Modelling\\
  F. Hoffmann-La Roche Ltd.\\
  Basel, Switzerland \\
}

\begin{document}
% \linenumbers
\maketitle
\begin{abstract}
Pharmaceutical research and development has accumulated vast and heterogeneous archives of data. Much of this knowledge stems from discontinued programs, and reusing these archives is invaluable for reverse translation. However, in practice, such reuse is often infeasible. In this work, we introduce \textit{DiscoVerse}, a multi-agent co-scientist designed to support pharmaceutical research and development at Roche. Designed as a human-in-the-loop assistant, \textit{DiscoVerse} enables domain-specific queries by delivering evidence-based answers: it retrieves relevant data, links across documents, summarises key findings and preserves institutional memory. 
% Given that automated evaluation metrics are poorly aligned with scientific utility, w
We assess \textit{DiscoVerse} through expert evaluation of source-linked outputs. Our evaluation spans a selected subset of 180 molecules from Roche's research and development repositories, encompassing over 0.87 billion BPE~\cite{sennrich-etal-2016-neural} tokens and more than four decades of research. To our knowledge, this represents the first agentic framework to be systematically assessed on real pharmaceutical data for reverse translation, enabled by authorized access to confidential archives covering the full lifecycle of drug development. Our contributions include: role-specialized agent designs aligned with scientist workflows; human-in-the-loop support for reverse translation; expert evaluation; and a large-scale demonstration showing promising decision-making insights. In brief, across seven benchmark queries, \textit{DiscoVerse} achieved near-perfect recall ($\geq 0.99$) with moderate precision ($0.71-0.91$). Qualitative assessments and three real-world pharmaceutical use cases further showed faithful, source-linked synthesis across preclinical and clinical evidence.
\end{abstract}

%=========================================================================================
\section{Introduction}
The pharmaceutical industry has generated vast repositories of experimental data over decades. For each drug candidate, companies typically produce extensive internal study reports, raw experimental data, toxicological findings, histopathological evaluations, nonclinical and clinical study presentations, and documentation of decision-making processes and content throughout drug discovery and development. Collectively, these materials represent an invaluable yet underutilized resource. Embedded within them are insights into treatment-related findings, target organ toxicities, information on the safety and efficacy profiles of the drug candidates as well as detailed experimental methodologies.

A significant portion of this archived information originates from discontinued drug candidates halted at various stages of development. Such attrition is inherent to pharmaceutical research and development, yet these data are far from failures: through \textbf{reverse translation}~\cite{kasichayanula2018reverse, honkala2022harnessing, vanmeerbeek2023reverse, sokol2025artificial}, insights from clinical outcomes can refine earlier‑stage research, turning therapeutic failures into opportunities to identify new targets and biomarkers~\cite{shakhnovich2018s} and to apply lessons from clinical pharmacology and regulatory reviews to new therapies~\cite{faucette2018reverse}.  A good example of such reverse translation efforts is discovery of a new therapeutic target (the TGF‑$\beta$ pathway~\cite{mariathasan2018tgfbeta}) while exploring clinical observations of treatment resistance and tracing them back to the underlying biology that can be therapeutically modulated. A challenge, however, in this reverse translation exercise is that end‑to‑end drug development produces thousands of heterogeneous documents dispersed across large organizations and external partners, with inconsistent naming conventions and project‑specific terminology that evolves over time. This fragmentation makes systematic retrieval via browsing or simple keyword search infeasible. As a result, synonymy and nomenclature drift, facts locked in tables, and context‑dependent phrasing yield high false negatives (missing evidence) and high false positives (noise). Overcoming these barriers requires \textbf{semantic retrieval}, \textbf{cross‑document linking}, \textbf{preclinical–clinical data alignment} and \textbf{auditable synthesis}~\cite{honkala2022harnessing, henrique2025data}. When curated, structured, and made searchable, these legacy datasets become reusable assets for quantitative analysis, machine learning, and predictive toxicology, enabling cross‑program pattern recognition across molecular structures, pharmacological targets, toxicological profiles, and safety margins, and ultimately improving decision‑making while reducing redundant experimentation.

Recent large language models (LLMs)~\cite{achiam2023gpt,team2023gemini,claude3,bai2023qwen,liu2024deepseek,guo2025deepseek} can read and synthesize massive unstructured corpora and, when domain‑adapted, generate concise, source‑grounded summaries of complex biomedical and medical text~\cite{tung2024comparison,van2024adapted,williams2025physician,ghim2023transforming,wornow2025zero,gallifant2025tripod,lai2024assessing}. This makes LLMs a practical way to drive insights across programs, applying consistent criteria over many reports so evidence that humans might miss still surfaces.

However, pharmaceutical research and development presents unique challenges that expose the limitations of single-agent LLM approaches. It is a highly specialized domain in which each study is context-specific with differences in endpoints, assays, and terminologies, making it difficult to provide standardized instructions or representative examples for LLMs to learn from. Moreover, large pharmaceutical companies often collaborate with multiple external contract research organizations (CROs) to optimize distribution of internal capabilities, costs, and speed. While these distributed processes enhance productivity, CROs also introduce additional heterogeneity. Consequently, single-agent LLMs often struggle to generalize robustly from sparse or non‑representative sources of information~\cite{pmlr-v235-reizinger24a, sengupta-etal-2025-exploring, peters2025generalization}. These challenges are further compounded by siloed data sources and the gradual drift of domain-specific terminology over time, which collectively hinder model consistency and reliability.

We therefore adopt a multi‑agent system (MAS)~\cite{boiko2023emergent, hong2023metagpt, wu2024ehrflow, kapoor2024ai, wang2025colacare, gottweis2025towards, zhu2025healthflow} architecture with role‑specialized agents for retrieval, research, verification, and synthesis. By enabling agents to interact and critique one another and enforcing source‑grounded outputs, MAS better handles sparse demonstrations while maintaining traceable, auditable reasoning chains, which is an essential requirement in regulated pharmaceutical research. Emerging empirical evidence further substantiates the advantages of MAS over single-agent LLMs, particularly in specialized domains with limited instructional examples~\cite{wang-etal-2024-rethinking-bounds}. For instance, Chen et al.~\cite{chen2025enhancing} report that MAS exhibits enhanced diagnostic capabilities relative to individual LLMs in clinical domains. Recent controlled evaluations~\cite{wang2025colacare,jin2025stella,zhu2025healthflow,wu2024ehrflow,huang2025biomni, zhu2025medagentboard} further demonstrate that appropriately designed MAS can achieve human-level performance in complex medical tasks, underscoring the viability of this approach for pharmaceutical research and development~\cite{seal2025aiagentsdrugdiscovery}.

Building on MAS framework, we developed \textit{DiscoVerse}, a multi-agent co-scientist tailored specifically for pharmaceutical knowledge ingestion and translational support across the drug development lifecycle. The system integrates and aligns preclinical and clinical data, enabling systematic preclinical–clinical mapping and comparison. Through its role-specialized agents for retrieval, research, validation, and synthesis, \textit{DiscoVerse} can efficiently uncover relevant information from historical archives to link preclinical observations with clinical outcomes. In design, \textit{DiscoVerse} functions as a human-in-the-loop assistant: project teams can query it with questions such as “What was the main reason for discontinuation of drug X?”. By retrieving and connecting molecular, toxicological, and clinical evidence, \textit{DiscoVerse} supports reverse translation, allowing scientists to derive meaningful insights from past clinical experiences and apply them to current research programs. 

The key contributions of our framework are four-fold: (i) \textbf{specialised agent designs} that mirror scientist workflows, each optimised for a distinct sub-task with external knowledge available; (ii) integration with \textbf{human-in-the-loop into the drug-design workflow}, moving from long-document ingestion through domain-specific question-answering and comparative historical analysis to actionable design insights and reverse translation; (iii) \textbf{expert evaluation on real-world pharmaceutical data}: unlike prior work that relies primarily on curated benchmark datasets, we build and systematically evaluate our framework using authentic pharmaceutical research and development data; and (iv) \textbf{large-scale demonstration} on a real pharmaceutical knowledge repository, showing promising results in answer accuracy and decision-context insights that were previously time-consuming to assemble.

By deploying a tailored multi-agent pharmaceutical co-scientist system, we aim to preserve expert knowledge, reduce human bias, and make information easily accessible. \textit{DiscoVerse} consolidates data from multiple sources, and efficiently extracts relevant insights from vast amounts of information. This ensures that critical knowledge is not only preserved when scientists leave or move on, but is also readily available and objectively presented to inform current drug-development decisions. Through this approach, \textit{DiscoVerse} addresses the dual challenge of knowledge preservation and accessibility, combining cutting-edge LLM capabilities with the practical needs of pharmaceutical research and development teams.

%=========================================================================================
\section{Methodology}

\begin{figure}[h]
    \centering
    \includegraphics[width=\linewidth]{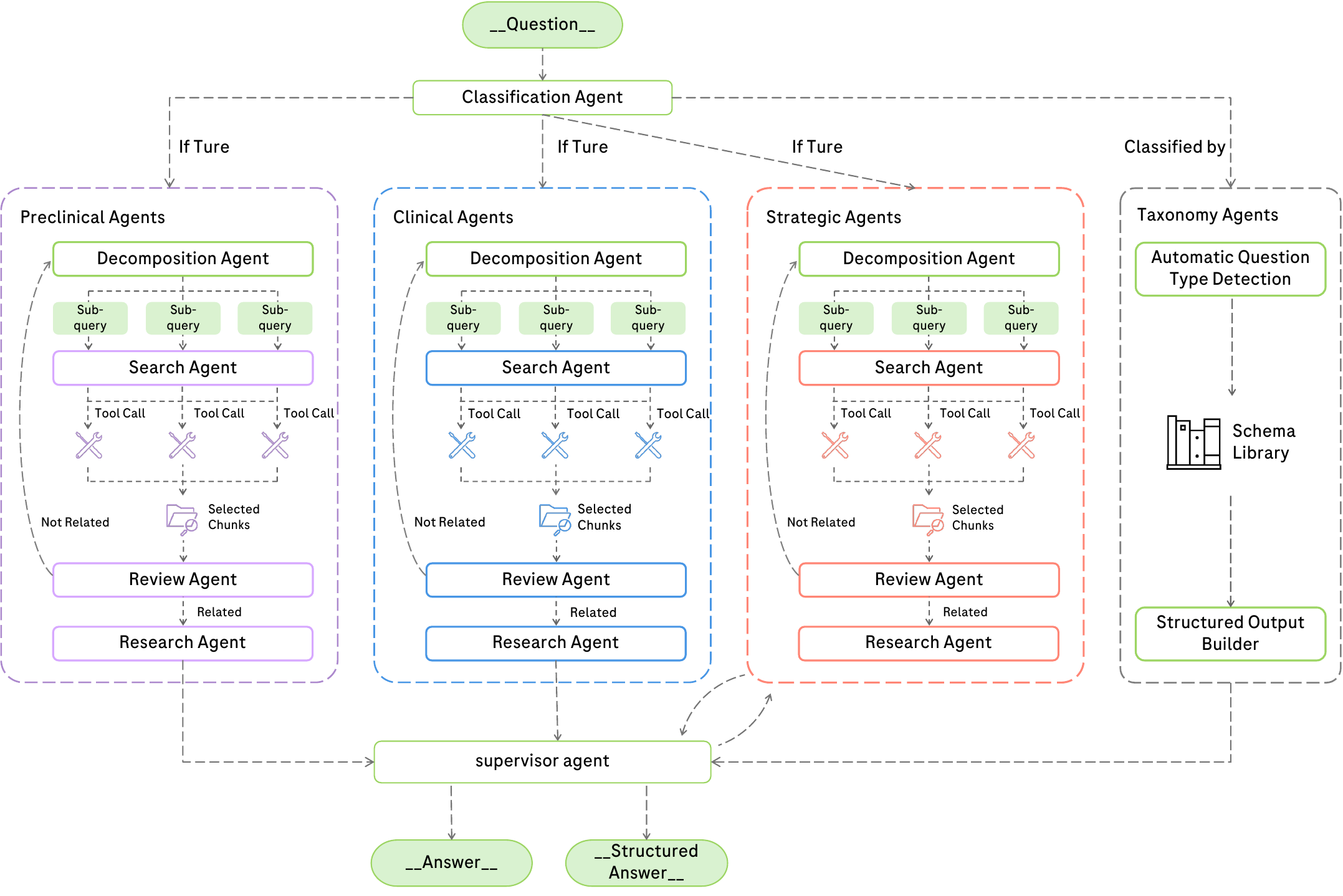}
    \caption{The illustrative overview of \textit{DiscoVerse}. The Discoverse orchestrates three specialized \textbf{Preclinical}, \textbf{Clinical}, and \textbf{Strategic} branches. Within each branch, \textbf{Decomposition Agent} follows a read–rewrite–retrieve~\cite{ma-etal-2023-query} workflow by generating sub-queries, calling tools to access the DiscoVerse database, and retrieving relavant chunks. If gaps remain, \textbf{Review Agent} will return to decomposition agent for refinement. A \textbf{Research Agent} discovers and synthesizes scientific findings, while the \textbf{Supervise Agent} orchestrates the process, deciding which branch outputs to integrate. Finally, a \textbf{Taxonomy Agent} produces structured answers through an expert-in-the-loop schema library for consistent answer output.}
    \label{fig:framework}
\end{figure}

\subsection{System Architecture and Agents}

\textit{DiscoVerse} is implemented as a modular multi-agent system with retrieval tools calling, as shown in Fig.~\ref{fig:framework}. Each agent in the system has a specialized role in the overall pipeline:

\begin{enumerate}
    \item \textbf{Preclinical Agents} handle queries related to research happening before entry into human in in-vitro and animal studies (e.g. in vitro safety assays, rodent/non-rodent repeated dose toxicity studies etc).
    \item \textbf{Clinical Agents} handle queries related to research activities happening after entry into human (e.g. clinical trials).
    \item \textbf{Strategic Agents} focus on higher-level questions, such as strategic reasons for portfolio decisions, or cross-project comparisons.
\end{enumerate}

These three agents are overseen by a \textbf{Supervisor Agent} that orchestrates the flow of the conversation and integrates findings from multiple domain agents. The multi-agent design allows each agent to use domain-specific prompts and logic, analogous to the coordinated work of a team of collaborating scientists. All agents communicate via structured messages, and the Supervisor ensures that their contributions are merged correctly. 

Moreover, to ensure consistent structured output~\cite{karmaker-santu-feng-2023-teler, dagdelen2024structured, zaratiana-etal-2025-gliner2}, we introduce \textbf{Taxonomy Agents} that operate downstream of the domain agents. They take the evidence and summaries produced by the Preclinical, Clinical, and Strategic Agents and map them into a schema library. The schema library is built collaboratively with scientists and project leads, and encodes expert insights into predefined question types. Each question type contains detailed descriptions, classification logic, and schema for structured output. For every query type, the schema library defines how an agent should identify intent, extract key entities or evidence, and assemble results into auditable summaries. Importantly, a single user query can map to multiple categories, allowing flexible composition across preclinical, clinical, and strategic dimensions.

\subsection{Agent Workflow}
When a user query is submitted to \textit{DiscoVerse}, it undergoes a series of processing steps involving the agents described above. The end-to-end workflow is outlined below:

\textbf{Query Classification and Decomposition (\textit{Classification and Decomposition Agent} in Fig.~\ref{fig:framework})} The classification agent analyzes the user's query and determine which knowledge domains it related to (preclinical, clinical, strategic, or a combination). The system then decomposes them into simpler but more detailed sub-queries using predefined decomposition rules. Simultaneously, the classification agent classifies the query into specific question types (such as toxicity, efficacy, or discontinuation reasons) based on categories defined collaboratively by project leaders and scientists. The \textit{DiscoVerse} maintains a library of structured output schemas (templates) corresponding to each question type, and when a query spans multiple topics, the system combines relevant schemas into a single composite template to ensure the final answer captures all necessary facets.

\textbf{Document Retrieval (\textit{Search Agent} in Fig.~\ref{fig:framework})} Once the sub-queries are defined and assigned to the appropriate domain agent, the next step is to retrieve relevant information from database. We employ a Hybrid document retriever that combines symbolic search with semantic search. The detailed methods are provided in Appendix~\ref{rag}.

\textbf{Relevance Reviewing (\textit{Review Agent} in Fig.~\ref{fig:framework})} The retrieval step typically yields a collection of document chunks that potentially contain the answer. Not all retrieved text will be truly relevant or high-quality, so each domain agent includes a review sub-process that filters these chunks. We utilize a reranker and an LLM-based relevance scoring mechanism: for each retrieved chunk, the agent’s LLM is prompted with a question-answering relevance check. The reranker returns a relevance score which is mapped to a float value in $[0,1]$ and LLM returns a judgment. Only chunks that both exceed a reranker score of 0.7 and have a relevance judgment are retained for analysis; the remainder are discarded to reduce noise.

\textbf{Domain-Specific Evidence Extraction (\textit{Research Agent} in Fig.~\ref{fig:framework})} After filtering, each domain agent now has a curated set of relevant text passages for its sub-query. The next step is evidence extraction and synthesis. The agent composes a prompt that includes the user’s sub-query and the collected text chunks (often truncated or summarized if they are long) and asks the LLM to generate a concise finding. This is effectively a summary or answer drawn from the evidence, focusing on the key details that address the question. The prompt templates for this step are tailored to each domain. We refer to these outputs as domain-specific findings. Throughout this process, the agents maintain the lineage of information so that evidence traceability is preserved (useful for later expert review, although in this paper we focus on the methodology rather than the user interface).

\textbf{Multi-Agent Synthesis (\textit{Supervisor Agent} in Fig.~\ref{fig:framework})} The Supervisor (or orchestrator) agent waits for the domain agents to finish their analysis. Once all requested domain-specific findings are ready, the Supervisor integrates the results and assembles a coherent answer that covers each aspect. It may simply concatenate the findings under appropriate subheadings or, if needed, generate a brief narrative that connects them. The important point is that each piece of the answer is coming from a specialized analysis pipeline, ensuring depth in that area, and the Supervisor merges these pieces into a comprehensive response to the original query. The multi-agent coordination is implemented such that if one domain yields “no finding” (for instance, if nothing relevant was found in clinical data), the system can still return the findings from other domains along with a note that no information was available in that domain. This ensures graceful handling of negative or null results.

\textbf{Consistent Structured Output (\textit{Taxonomy Agent} in Fig.~\ref{fig:framework})} Besides providing a long free-text report, \textit{DiscoVerse} outputs the answer in a structured form. Each field in the schema corresponds to a specific piece of information, as shown later in Fig.~\ref{fig:discontinue_text_sample} and \ref{fig:toxicity}. The content generated by the domain agents is mapped into these schema fields. If the query was multi-type and a composite schema was created, all relevant fields from each domain are included. In addition, the \textit{Taxonomy Agents} operationalize a modular \textit{schema library} co-designed with scientists and project leads, which contains predefined question types with concise descriptions, routing/classification rules, required evidence elements to extract, and structured output templates; a single user query may match multiple types, which are composed into a unified schema while preserving per-type provenance to support auditable synthesis.

Throughout this pipeline, by breaking the problem into smaller tasks handled by different agents, we reduce the cognitive load on any single LLM prompt and make the overall process more interpretable. The modular design also allowed us to incorporate fallback mechanisms: for example, within Classification Agent, if the LLM classification fails or is not available, a rule-based classifier steps in to ensure the query is still routed properly. Similarly, if the retriever finds too much data, the system can impose limits or stricter relevance thresholds to keep only the most salient information. We aim to continuously ingest and analyze historical research and development knowledge, and to provide scientists and decision-makers with an ever-ready, unbiased assistant to inform their next steps.

%=========================================================================================
\section{Materials}
\subsection{Data Selection and Description}
Our documentation repository contains 872,453,585 BPE~\cite{sennrich-etal-2016-neural} tokens~\footnote{We computed token counts using OpenAI’s tiktoken BPE tokenizer~(\url{https://github.com/openai/tiktoken}) on text extracted from PDFs.} in 15,762 PDF files covering 180 selected drug molecules. On average, each molecule is associated with approximately 10,000 pages of documentation, forming a comprehensive collection that includes animal study reports, clinical study reports, meeting minutes, project summary presentations, and investigator brochures. A substantial fraction of the data originates from pre-clinical \textit{in vivo} animal studies.

The dataset spans a period of more than four decades. Consequently, the collection includes scanned copies of legacy documents, exhibiting considerable variability in linguistic style, level of technical detail, and structural format. This heterogeneity, particularly the presence of non-digitally born documents, presents a substantial challenge for reliable text retrieval, parsing, and data harmonization.

\subsection{Data Preparation}
To organize this dataset, all documents were grouped according to the internal unique identifier corresponding to each molecule. Documents referring to multiple molecules (e.g., combined toxicity or efficacy studies, portfolio overview reports, and project closure presentations) were assigned to all relevant identifiers to preserve completeness. The detailed data preparation pipeline (including embedding and indexing) is described in Appendix~\ref{rag}.

%=========================================================================================
\section{Evaluation Design}

Evaluating LLMs in pharmaceutical applications requires particular care given the safety-critical nature of drug development decisions. A key concern is hallucination, which poses significant risks when LLMs are deployed to extract clinical data or support regulatory submissions~\cite{hakim2025need}. However, standard evaluation approaches using automated metrics like ROUGE~\cite{lin-2004-rouge} and BLEU~\cite{papineni-etal-2002-bleu} are insufficient for this domain, as they correlate poorly (sometimes negatively) with how medical experts judge accuracy~\cite{wang-etal-2023-automated, ben-abacha-etal-2023-investigation,  nguyen2024comparative, fraile2025expert}. Expert evaluation becomes essential and crucial for medical and drug development applications~\cite{agrawal2025evaluation, kim2025ragevaluation, schilling2025text, zhu2025medagentboard, rieff2025smmile, gu2025medagentaudit}. We therefore use expert evaluation of source-linked outputs; while this increases review burden, it better captures scientific utility. Our evaluation protocol, centered on expert judgment, distinguishes \textit{DiscoVerse}, and to our knowledge no prior research has systematically evaluated agentic systems for drug discovery workflows on real pharmaceutical data, particularly for reverse translation. Additionally, \textit{DiscoVerse} is the first platform with authorized access to confidential data generated during the drug development process. Such access is essential for the nature of our research, which requires comprehensive and context-rich information spanning the full drug development lifecycle.

\subsection{Question Design}
To ensure our evaluation reflects authentic scientist needs and comprehensively tests \textit{DiscoVerse}'s capabilities, we collaborated actively with toxicology project leaders and safety clinical and pre-clinical scientists to identify the types of questions routinely asked when evaluating discontinued programs. Therefore, we designed nine \textbf{preclinical and clinically relevant questions spanning the entire drug development lifecycle}, each intended to probe a distinct facet of information extraction relevant to late-stage drug development. These questions were derived from real decision-making scenarios and needs: assessing whether a shelved compound could be repurposed, understanding why similar molecular scaffolds failed previously, or determining the optimal upper dose range for new candidate dose-range-finding studies using historical data and precedent analyses explaining prior dosing decisions. Collectively, these queries represent a spectrum of complexity, ranging from simple factual retrieval to complex synthesis of disparate information. Meanwhile, the design strategy ensures that our evaluation is both \textbf{scientifically grounded} (reflecting questions scientists actually need answered) and \textbf{comprehensively diagnostic} (systematically testing each capability required for effective pharmaceutical knowledge extraction).

\begin{custommdframed}
    \textbf{Simple Factual Retrieval}: These queries test the agent's core ability to locate a single, explicit data point embedded within lengthy technical documents. \\
    \\
    - \textbf{Q1: First-in-Human (FIH) Dose}: "What was the first in human dose for drug X?"
    
    - \textbf{Q2: Route of Administration (RoA)}: "What was the route of administration in humans for drug X?" 
    \end{custommdframed}
    
    \begin{custommdframed}
    \textbf{Retrieval with Comparative Reasoning}: This query requires the agent to not only find multiple data points across dose-escalation studies but also perform a comparison to identify a specific value (e.g., the maximum tolerated dose). \\
    \\
    - \textbf{Q3: Highest Dose in Phase I Multiple Ascending Dose (MAD) or Phase II studies}: "What was the highest clinical dose in Phase I (MAD)/II for drug X?"
\end{custommdframed}

\begin{custommdframed}
    \textbf{Relational Extraction}: This task tests the agent's ability to identify and validate the relationship between two distinct entities (e.g., a specific dose level and a severe adverse event), which is a known challenge for LLMs that often struggle with precise entity linking across document sections. \\
    \\
    - \textbf{Q4: Highest Dose with Severe Adverse Events (SAEs)}: "What was the highest clinical dose at which there were severe adverse events for drug X?"
\end{custommdframed}

\begin{custommdframed}
    \textbf{Contextual Synthesis}: These are the most complex queries, requiring the agent to find multiple pieces of information from potentially different sections (or even different documents) and synthesize them into a coherent, contextually accurate summary that captures implicit expert reasoning. \\
    \\
    - \textbf{Q5: Efficacious Dose}: "What was the efficacious dose in the clinic?" (requires integrating efficacy endpoints, biomarker data, and dose-response relationships).
    
    - \textbf{Q6: Treatment Regimen}: "What was the treatment regimen for drug X in humans?" (demands synthesis of dosing frequency, duration, and administration context).
    
    - \textbf{Q7: Margin of Safety}: "What do we know about the Margin of Safety of drug X?" This query is particularly challenging as it may require synthesizing preclinical safety data (e.g., NOAEL from toxicology studies) with clinical dose information to describe a derived concept not explicitly stated in the text, which is mimicking the integrative reasoning pharmaceutical scientists perform when assessing compound viability.
\end{custommdframed}

\begin{custommdframed}
    \textbf{Strategic Decision Context}: These queries test the agent's ability to extract high-level strategic information that informed program termination decisions, which is critical for understanding institutional lessons learned.

    - \textbf{Q8: Discontinuation Rationale}: "What primary reason led to the discontinuation of drug X, and in which development phase was this decision made?" This query requires identifying decision-making context that may be scattered across preclinical, clinical phase, and strategic decisions, then distilling it into actionable insights.
\end{custommdframed}

\begin{custommdframed}
    \textbf{Multi-Phase Evidence Integration with Structured Output}: This most challenging query tests the agent's ability to perform complex cross-document synthesis while adhering to specific output formatting requirements.

    - \textbf{Q9: Multi-Phase Toxicity Evidence Integration}: "Is drug X hematotoxic and if yes, in which animal species and in humans? What is the evidence that led to this conclusion? Please split the evidence into pre-clinical and clinical and if there is evidence in both phases then list both pieces of evidence one after the other." 
    
    This query demands: (i) binary classification (hematotoxic yes/no), (ii) species-specific identification across preclinical models, (iii) human clinical evidence extraction, (iv) evidence categorization and structured presentation, and (v) integration of findings across the entire drug development lifecycle, which mimics the comprehensive toxicological assessments scientists perform when evaluating compound safety profiles.
\end{custommdframed}

\subsection{Evaluation Design}

\begin{table}[hptb]
    \centering
    \small
    \setlength{\tabcolsep}{3pt}
    \renewcommand{\arraystretch}{1.25}
    \resizebox{\textwidth}{!}{
    \begin{tabular}{@{}lp{3.5cm}p{2.5cm}p{3.5cm}p{2.5cm}@{}}
    \toprule
    & \textbf{TP} & \textbf{TN} & \textbf{FP} & \textbf{FN} \\
    \midrule
    \textbf{Q1: FIH Dose} &
    Value Identified \cmark{} AND Context (if given) \cmark{} &
    Correctly states info is absent \cmark{} &
    Incorrect Value \xmark{} OR Preclinical Value \xmark{} OR Hallucination \xmark{} &
    Fails to find present value \xmark{} \\
    \addlinespace
    \textbf{Q2: RoA} &
    Correct Route \cmark{} AND Context (if given) \cmark{} &
    Correctly states info is absent \cmark{} &
    Incorrect Route \xmark{} OR Preclinical Route \xmark{} &
    Fails to find present RoA \xmark{} \\
    \addlinespace
    \textbf{Q3: Highest Dose} &
    Highest Value Identified \cmark{} AND Correct Phase \cmark{} &
    Correctly states info is absent \cmark{} &
    Incorrect Value \xmark{} OR Incorrect Phase \xmark{} OR Not Highest Value \xmark{} &
    Fails to find present value \xmark{} \\
    \addlinespace
    \textbf{Q4: Dose w/ SAEs} &
    Correct Dose \cmark{} AND Correct Link to SAE \cmark{} &
    Correctly states info is absent \cmark{} &
    Incorrect Dose \xmark{} OR Incorrect Link to SAE \xmark{} OR Confuses AE/SAE \xmark{} &
    Fails to connect present dose--SAE link \xmark{} \\
    \addlinespace
    \textbf{Q5: Efficacious Dose} &
    Correct Dose \cmark{} AND Context (if given) \cmark{} &
    Correctly states info is absent \cmark{} &
    Incorrect Dose \xmark{} OR Misrepresents Efficacy \xmark{} &
    Fails to find present efficacy info \xmark{} \\
    \addlinespace
    \textbf{Q6: Regimen} &
    All components correct (Dose, Freq., Duration) \cmark{} &
    Correctly states info is absent \cmark{} &
    One or more components incorrect \xmark{} &
    Fails to synthesize present regimen details \xmark{} \\
    \addlinespace
    \textbf{Q7: Safety Margin} &
    Synthesizes correct preclinical AND/OR clinical data \cmark{} &
    Correctly states info is absent \cmark{} &
    Uses incorrect values (e.g., toxic dose) for synthesis \xmark{} &
    Fails to connect present data to describe margin \xmark{} \\
    \bottomrule
    \end{tabular}
    }
    \caption{Definition of TP, TN, FP, FN for seven benchmark queries respectively.}
    \label{tab:metrics}
\end{table}

To demonstrate the value \textit{DiscoVerse} brings to pharmaceutical decision-making, we design the evaluation targeting the core challenge faced by research and development teams: \textit{extracting discrete, decision-critical information from decades of unstructured, heterogeneous documentation}. This evaluation directly mirrors real-world scenarios where scientists must synthesize evidence across discontinued programs to inform current development decisions. This task traditionally requiring weeks of manual review or remaining unaddressed due to resource constraints.

To keep the study rigorous yet tractable, we selected seven representative (Q1 - Q7), decision-critical queries for quantitative expert-in-the-loop evaluation. For each query, \textit{DiscoVerse} retrieved source-linked document chunks and generated answers that pharmaceutical scientists independently adjudicated against the original sources, labeling each response as True Positive (TP), True Negative (TN), False Positive (FP), or False Negative (FN) according to prespecified criteria for factual accuracy and contextual completeness~\cite{gartlehner2024promise}. We choose \textbf{accuracy}, \textbf{precision}, \textbf{recall}, \textbf{specificity}, \textbf{F1-score} to evaluate the performance. Detailed evaluation criteria and per-query rubrics are provided in Tab.~\ref{tab:metrics} and Appendix~\ref{expert_eva} and~\ref{metrics}. The remaining two queries (Q8 and Q9) were not scored quantitatively across all drug programs because (i) the original project leads have left, and the underlying “real reason” is not written explicitly in our decades‑spanning repository; it is latent, spread across reports, raw data, figures, and protocol amendments, making consistent program-wide scoring infeasible; and (ii) some key archival documents are missing or incomplete. This is exactly why we built \textit{DiscoVerse}: role‑specialized agents retrieve, link, and synthesize dispersed evidence to surface these otherwise hidden answers. Even though the answers we provide for Q8 and Q9 cannot be evaluated quantitatively for the two reasons mentioned above, they hold significant value in offering an overview of the information found and available in the documents associated with the particular Roche molecule. These findings also provide valuable insights for further understanding and analysis.

%%%%%%%%%%%%%%%%%%%%%%%%%%%%%%%%%%%%%%%%%%%%%%%%%%%%%%%%%%%%
\section{Results}

\subsection{Experimental Setup}
Our experiments aim to test and assess the system-level performance of \textit{DiscoVerse}'s multi-agent architecture rather than benchmarking individual LLM capabilities. All agents were powered by GPT-4.1 (gpt-4.1-2025-04-14) to isolate architectural contributions from model-specific variations. We run the embedding and reranker model locally on NVIDIA A100 GPU.

We executed \textit{DiscoVerse} for each query (Q1–Q9) on each molecule in our dataset, generating both unstructured (natural language) and structured (formatted) outputs for each molecule-query pair. To enable rigorous expert evaluation and ensure traceability, we logged all retrieved document chunks, intermediate agent outputs, and source attributions.

\subsection{Quantitative Evaluation Results}

\begin{table}[htbp]
\centering
\caption{Per-query performance of \textit{DiscoVerse}}
\label{tab:query-metrics}
\begin{tabular}{cccccc}
\hline
Query & Accuracy & Precision & Recall & Specificity & F1-Score \\
\hline
Q1 & 0.9086 & 0.8712 & 1.0000 & 0.7605 & 0.9311 \\
Q2 & 0.8852 & 0.8444 & 1.0000 & 0.6956 & 0.9156 \\
Q3 & 0.8524 & 0.7890 & 1.0000 & 0.6707 & 0.8820 \\
Q4 & 0.9243 & 0.8488 & 0.9864 & 0.8828 & 0.9125 \\
Q5 & 0.8857 & 0.7142 & 1.0000 & 0.8400 & 0.8333 \\
Q6 & 0.9426 & 0.9078 & 1.0000 & 0.8679 & 0.9517 \\
Q7 & 0.8677 & 0.8139 & 1.0000 & 0.6862 & 0.8974 \\
\hline
\end{tabular}
\end{table}

We evaluated \textit{DiscoVerse} across seven benchmark queries designed to simulate real pharmaceutical information retrieval tasks. Complete metrics (accuracy, precision, recall, specificity, and F1-score) appear in the Tab.~\ref{tab:query-metrics}. The results show a consistent performance pattern that defines how the system should be used in drug development workflows.

Across queries, \textit{DiscoVerse} achieves near-perfect recall (1.0000 for six queries; 0.9864 for Q4), ensuring that relevant items are rarely missed. Where performance varies is in precision and specificity. Precision ranges from 0.7142 (Q5) to 0.9078 (Q6), indicating a tendency toward false positives. Complementing this view, specificity spans 0.6707 (Q3) to 0.8828 (Q4), showing how often the system correctly rejects non-matching content. In other words, when the agent admits borderline or context-mismatched items, both precision (positive-set purity) and specificity (negative-set rejection) decline. Lower specificity reflects many negatives being incorrectly marked as positives. Together with the precision range above, this paints a consistent picture: the system aggressively retrieves candidates (high recall) but sometimes over-includes near-miss evidence, lowering both precision and specificity.

Expert evaluation indicates the false positives are not hallucinations but failures of contextual understanding. Each error class simultaneously reduces precision (adds incorrect items to the positive set) and specificity (converts true negatives into false positives): the agent might extract a preclinical dose when asked for a clinical one (described in Tab.~\ref{tab:overall_error} for Q1 and Tab.~\ref{tab:overall_error} for Q3), describe planned as an actual administration route (described in Tab.~\ref{tab:overall_error} for Q1 and Tab.~\ref{tab:overall_error} for Q2), confuse trial phases (described in Tab.~\ref{tab:overall_error} for Q3), link a dose to an unrelated adverse event (described in Tab.~\ref{tab:overall_error} for Q4), report efficacy for doses planned but not yet tested in humans (described in Tab.~\ref{tab:overall_error} for Q5), provide incorrect dose-specific details like wrong infusion durations (described in Tab.~\ref{tab:overall_error} for Q6), or conflate preclinical safety margins with clinical data (described in Tab.~\ref{tab:overall_error} for Q7)

The performance profile with high recall and moderate precision suggests \textit{DiscoVerse} works best as an augmentation tool rather than a standalone system. It automates the most tedious part of knowledge work (comprehensive search) while leaving contextual verification to domain experts. This shifts the workflow from manual "search and synthesize" to AI-assisted "review and discover". 

\begin{table*}[htbp]
\centering
\caption{Error Analysis - Examples of False Positives and Correct Samples}
\label{tab:overall_error}
\scriptsize
\resizebox{\textwidth}{!}{
\begin{tabular}{lp{7cm}p{4cm}p{2.5cm}}%{XXXXXX}%{lccccc}
\toprule
\textbf{Correctness} & \textbf{Agent Answer} & \textbf{True Answer} & \textbf{Error Type} \\
\midrule
%----------Q1----------
\multicolumn{4}{l}{\textbf{Q1 (FIH Does)}}\\
\addlinespace
\quad \xmark{} - FP & The first-in-human dose for X was 100 mg/kg administered orally (by gavage) to cynomolgus monkeys in a single ascending dose study. & No planned dose in available data. Only first dosing available for monkeys. & Preclinical and clinical confusion \\
\addlinespace
\quad \xmark{} - FP & The first-in-human (FIH) dose for X was 0.75 mg/kg administered as a single oral dose to healthy volunteers. & The planned FIH was 0.75 mg/kg but has not yet been tested in humans. & Planned and actual confusion \\
\addlinespace
\quad \cmark{} - TP & \multicolumn{2}{p{10cm}}{The first-in-human dose for X was 400 mg as a single oral dose, administered to healthy Asian male subjects in Cohort 1 of the Phase I SAD study YP29017.} & ---\\
\addlinespace
\midrule
%----------Q2----------
\multicolumn{4}{l}{\textbf{Q2 (Route of Administration)}}\\
\addlinespace
\quad \xmark{} - FP & The route of administration in humans for drug X was oral (by gavage). & The dose was not yet tested in humans. The route was planned to be oral but not yet tested. & Planned and actual confusion \\
\addlinespace
\quad \cmark{} - TN & \multicolumn{2}{p{10cm}}{There is no evidence of X being administered to humans in any clinical trial according to the provided information. The oral route is repeatedly stated as the intended route but this is based solely on preclinical rationale.} & --- \\
\addlinespace
\quad \cmark{} - TP & \multicolumn{2}{p{10cm}}{In all human clinical trials, X was administered exclusively via the intranasal route using a metered-dose nasal spray device. No other routes of administration were used in humans} & --- \\
\addlinespace
\midrule
%----------Q3----------
\multicolumn{4}{l}{\textbf{Q3 (Highest Phase I/II Dose)}}\\
\addlinespace
\quad \xmark{} - FP & The highest clinical dose administered in Phase I (Multiple Ascending Dose) studies of X was $1.29 \times 10^{12}$ vector genomes (vg) per eye, delivered as a single intravitreal injection of 100 $\mu$L per eye. No Phase II clinical dose data are available. & No MAD studies were performed in humans. This dose was from a pre-clinical MAD study. & Preclinical and clinical confusion \\
\addlinespace
\quad \xmark{} - FP & The highest clinical dose administered in Phase I (Multiple Ascending Dose) trials for X was 40 mg as an intravenous (IV) infusion over 48 hours. Phase II studies had not been initiated or reported at the time of the latest available data. & The highest clinical dose (40 mg) was administered in Phase I Single Ascending Dose (SAD) trials, not Multiple Ascending Dose (MAD) trials. & SAD and MAD confusion \\
\addlinespace
\quad  \cmark{} - TP & \multicolumn{2}{p{10cm}}{The highest clinical dose of X administered in Phase I (MAD) trials was 35 mg once weekly (QW) intravenously, achieved in escalation cohorts (e.g., 20/20/35-mg). In Phase II trials, the highest dose tested was 15 mg QW or Q2W (in combination with atezolizumab), with most studies using 10 mg as the recommended dose for further development.} & ---\\
\addlinespace
\midrule
%----------Q4----------
\multicolumn{4}{l}{\textbf{Q4 (Highest Dose with SAEs)}}\\
\addlinespace
\quad \xmark{} - FP & The highest clinical dose at which SAEs were observed for X was 250 $\mu$g/kg administered intravenously. At this dose, SAEs included capillary leak syndrome (CLS, Grade 2 and Grade 4, the latter being life-threatening and not resolved at reporting), dyspnea (Grade 2), infusion-related reactions (Grade 2), and pyrexia (Grade 1/2, with one case requiring hospitalization). & At the 250 $\mu$g/kg dose, reported Grade 3+ SAEs included neutropenia, fatigue, peripheral neuropathy, thrombocytopenia, and anemia, which are not mentioned in the answer. & Incomplete SAE reporting \\
\addlinespace
\quad \xmark{} - FN & The highest clinical dose (SAEs, CTCAE Grade $\ge$3) were observed for X was 45 mg/kg once weekly (QW). However, none of these SAEs were considered related to X by the investigators, and no dose-limiting toxicities (DLTs) or drug-related SAEs were observed at any dose level. & At the 45 mg/kg dose, relevant SAEs like cachexia and pericardial effusion were reported but missed by the agent. & Missed relevant SAEs \\
\addlinespace
\quad \cmark{} - TP & \multicolumn{2}{p{10cm}}{The highest clinical dose at which severe adverse events (AEs) were reported for X was 2000 mg as a single oral dose in healthy volunteers. Severe AEs (flu-like symptoms: headache, pyrexia, nausea, myalgia, dizziness, malaise) occurred only at the 1600 mg and 2000 mg dose levels. These events were transient, managed with analgesics, and resolved within 1–2 days.} & --- \\
\addlinespace
\midrule
%----------Q5----------
\multicolumn{4}{l}{\textbf{Q5 (Efficacious Dose)}}\\
\addlinespace

\quad \xmark{} - FP & The highest dose of X tested in the clinic was 0.25 mg/kg IV. At this dose, stable disease was observed in 3 of 12 patients (25\%). Dose-limiting toxicities (capillary leak syndrome) occurred at 0.25 mg/kg, preventing further escalation. & Hallucination - capillary leak syndrome was not mentioned in the report. & Hallucinated adverse event \\
\addlinespace
\quad \xmark{} - FP & The efficacious dose of X in the clinic is 200–400 mg BID or 600–1000 mg QD. These doses demonstrated robust antiviral activity with median HBV DNA declines of 2.7–3.2 $\log_{10}$ IU/mL after 28 days. & Efficacy in humans has not yet been tested. & Preclinical and clinical confusion \\
\addlinespace
\quad \cmark{} - TP & \multicolumn{2}{p{10cm}}{The efficacious dose of X identified in clinical studies is 250 mg once weekly in combination with Y. This dose was selected as the recommended Phase 2 dose based on integrated PK, PD, safety, and efficacy data from Phase 1/2 trials.} & --- \\
\addlinespace
\midrule
%----------Q6----------
\multicolumn{4}{l}{\textbf{Q6 (Treatment Regimen)}}\\
\addlinespace

\quad \xmark{} - FP & X was administered as an IV infusion on Days 1, 3, and 5 of each 21-day cycle. Each infusion lasted 30 minutes, and treatment continued until disease progression or unacceptable toxicity. & Low dose infusion was 5 minutes. 30 minutes was for control or high dose only. & Incorrect dose-specific details \\
\addlinespace
\quad \xmark{} - FP & X was administered as IV infusion with weekly dosing at 20, 30, or 45 mg/kg. Infusion durations were reduced from 90 minutes for the first dose to 60 or 30 minutes for subsequent doses if tolerated. & Almost correct. Weekly dosing was up to 60 mg/kg, not 45 mg/kg. & Incorrect dose range \\
\addlinespace
\quad \cmark{} - TP & \multicolumn{2}{p{10cm}}{X was administered orally as immediate-release capsules. Regimens included single ascending doses (200–1600 mg), every other day dosing (50–2000 mg), and once weekly dosing (800/1200 mg). All regimens were observer- or double-blind with matching placebo controls.} & --- \\
\addlinespace
\midrule
%----------Q7----------
\multicolumn{4}{l}{\textbf{Q7 (Margin of Safety)}}\\
\addlinespace

\quad \xmark{} - FP & The Margin of Safety for drug X is considered adequate. In cynomolgus monkeys, the NOAEL (50 mg/kg/day) provides a 5.5-fold (AUC) and 6.2-fold (Cmax) margin over human exposure. In rats, margins are below unity (AUC margin <1x, Cmax ~1x). & NOAEL in rats was 150 mg/kg with margins of 16x and 97x to human Cmax, not below unity. & Incorrect margin calculations \\
\addlinespace
\quad \xmark{} - FP & The Margin of Safety for X is based on preclinical studies and initial Phase 1 clinical studies in healthy volunteers. In clinical studies, no dose-limiting toxicities were observed up to 100 mg single dose and 50 mg multiple dose. & X was not tested in humans yet. All data referenced is preclinical only. & Preclinical and clinical confusion \\
\addlinespace
\quad \cmark{} - TP & \multicolumn{2}{p{10cm}}{The Margin of Safety for X is established based on preclinical studies in cynomolgus monkeys and rats. The NOAEL in monkeys is 0.5 mg/kg/dose IV every 3 weeks, with adverse effects observed at $\ge$1.4 mg/kg/dose. No human clinical data are available in the first IB.} & --- \\
\bottomrule

\end{tabular}}
\end{table*}

\subsection{Qualitative Evaluation Findings}

Queries Q8 (Discontinuation Rationale) and Q9 (Multi-Phase Toxicity Evidence Integration) were evaluated qualitatively rather than through the quantitative TP/FP/TN/FN framework applied to Q1-Q7. In parallel, we continue collaborating with project leads to retrieve missing records and refine our approach through iterative feedback on prompt formulation and task decomposition. While quantitative evaluation of these responses is not feasible for the reasons noted above, they nonetheless provide a valuable synthesis of the available evidence across documents linked to each Roche molecule. This synthesis offers meaningful perspectives that enhance contextual understanding and inform subsequent methodological enhancements.
\paragraph{Discontinuation Rationale (Q8)} Q8 presents unique evaluation challenges because discontinuation of a drug candidate is typically reported under one or two official categories (e.g., clinical safety, preclinical safety, or clinical efficacy). In reality, however, discontinuation decisions are often multifactorial and result from the interaction of several contributing elements rather than a single definitive cause. A program may be terminated due to a combination of preclinical or clinical issues (e.g., safety, efficacy, or bioavailability), outcomes of clinical futility analyses, strategic portfolio reprioritization, and resource limitations. The relative influence of these factors reflects an institutional decision-making context that can be definitively confirmed only by the original project leads. Consequently, although the official documentation or labeling may list a single primary reason for discontinuation, such classifications rarely capture the full complexity of the underlying decision process.

\begin{figure}[t]
    \centering
    \includegraphics[width=\linewidth]{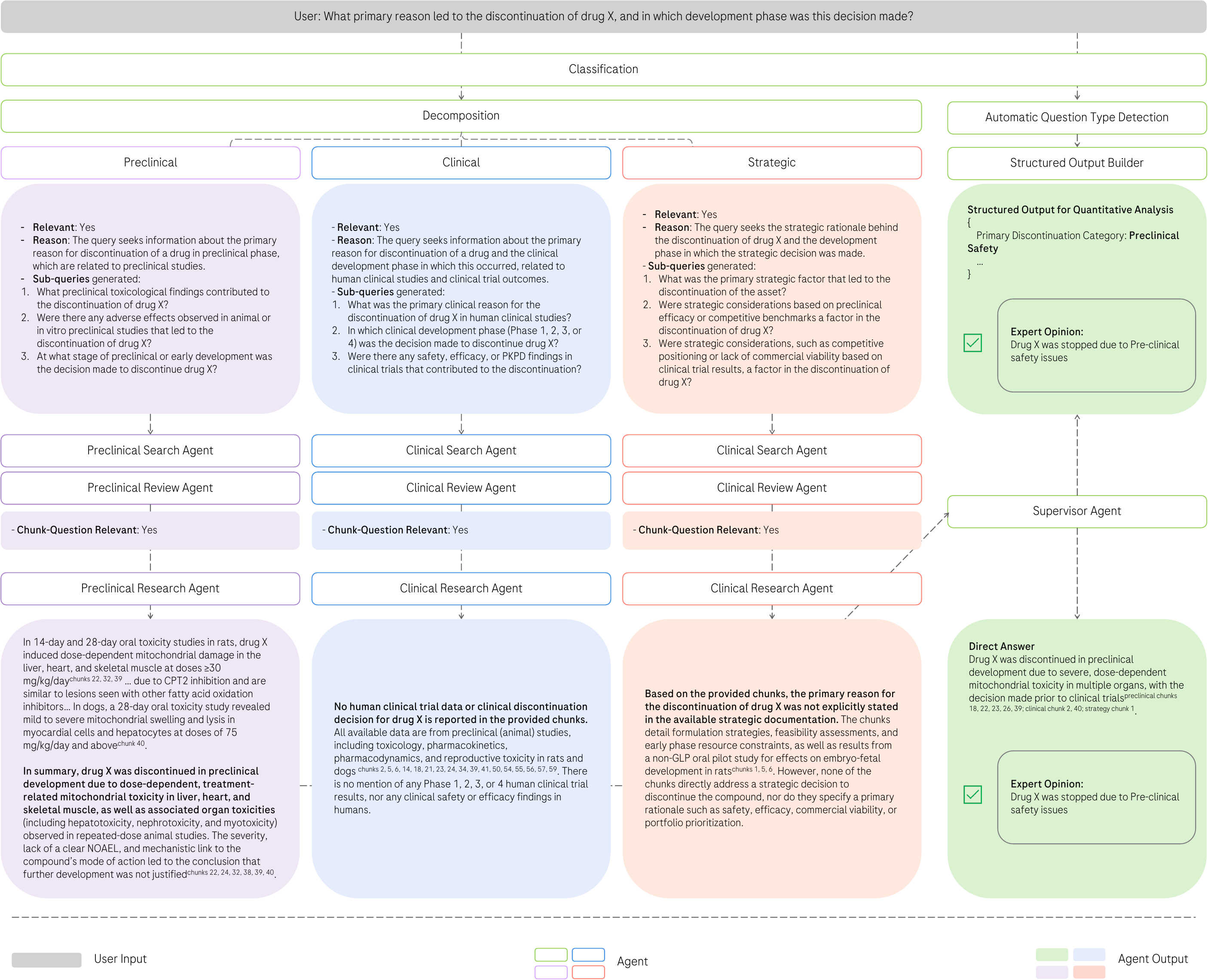}
    \caption{The illustrative output of Q8 (Discontinuation Rationale).}
    \label{fig:discontinue_text_sample}
\end{figure}

As shown in Fig.~\ref{fig:discontinue_text_sample}, this example illustrates how DiscoVerse can accurately identify that the reason for discontinuation originates in the preclinical phase, rather than being driven by clinical or other strategic factors.

\begin{figure}[t]
    \centering
    \includegraphics[width=\linewidth]{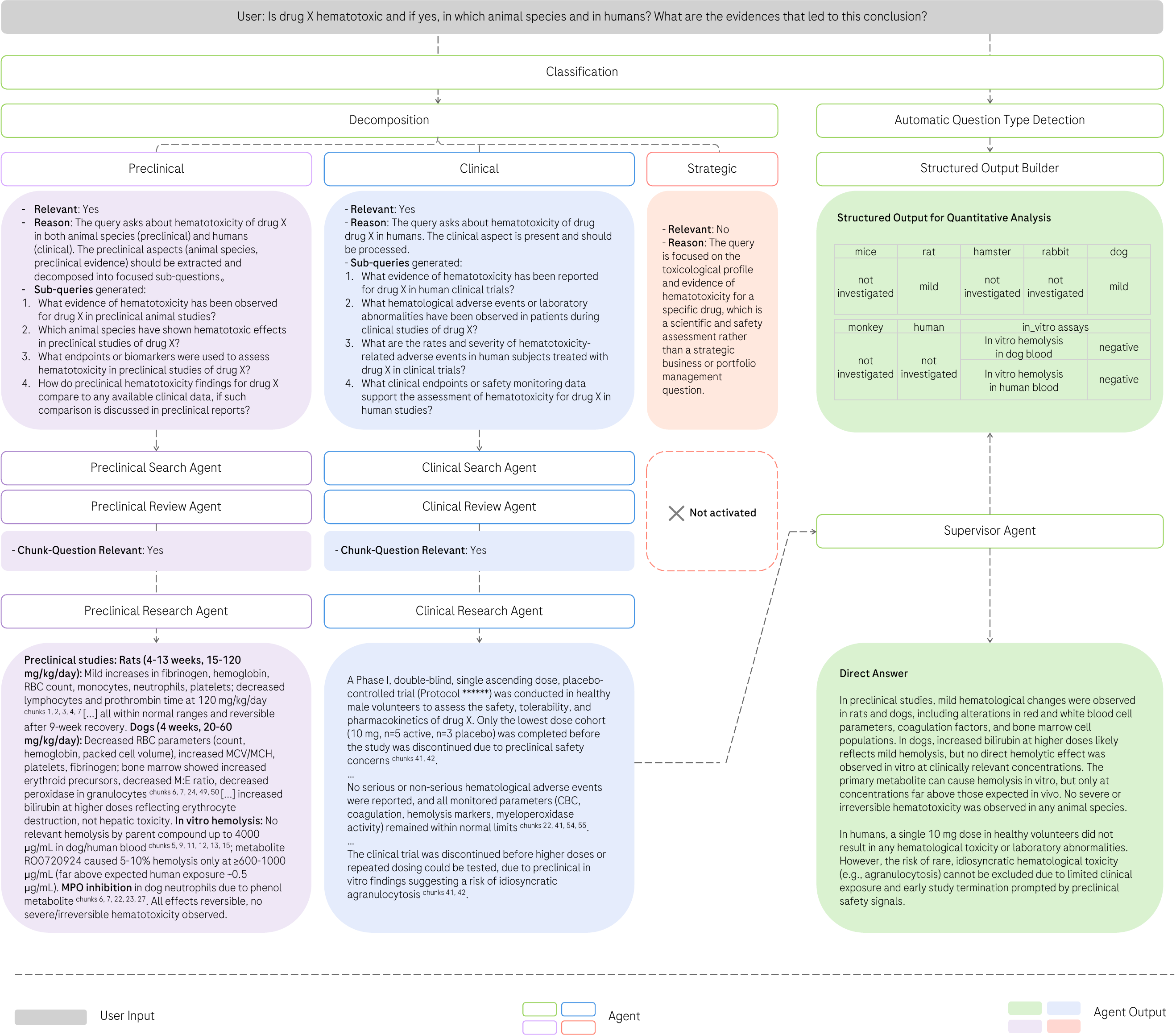}
    \caption{The illustrative output of Q9 (Multi-Phase Toxicity Evidence Integration).}
    \label{fig:toxicity}
\end{figure}

\paragraph{Multi-Phase Toxicity Evidence Integration (Q9)} Q9 presents analogous evaluation challenges: toxicity information is fragmented across the entire drug development lifecycle, spanning preclinical and clinical phases. Crucially, many of these documents contain evolving, preliminary, or context-dependent findings rather than definitive conclusions about whether a compound exhibits organ-specific toxicity (e.g., hematotoxicity). As visualized in Fig.~\ref{fig:toxicity}, \textit{DiscoVerse} decomposes the user query into distinct sub-tasks that separately interrogate preclinical and clinical evidence sources. The preclinical reasoning chain aggregates findings from 14-day and 28-day toxicity studies in rats and dogs, along with in vitro assays. Simultaneously, the clinical reasoning chain confirms that no human data are available, thereby preventing spurious claims of toxicity. The structured output builder consolidates these threads into a cross-species toxicity matrix summarizing the evidence base.

This example highlights the model’s ability to perform multi-phase evidence integration: it synthesizes distributed, heterogeneous data sources and reconciles apparently conflicting or incomplete evidence streams into a coherent structured summary.

\subsection{Real-World Pharmaceutical Industry Use Cases}

\begin{figure}[t]
    \centering
    \includegraphics[width=\linewidth]{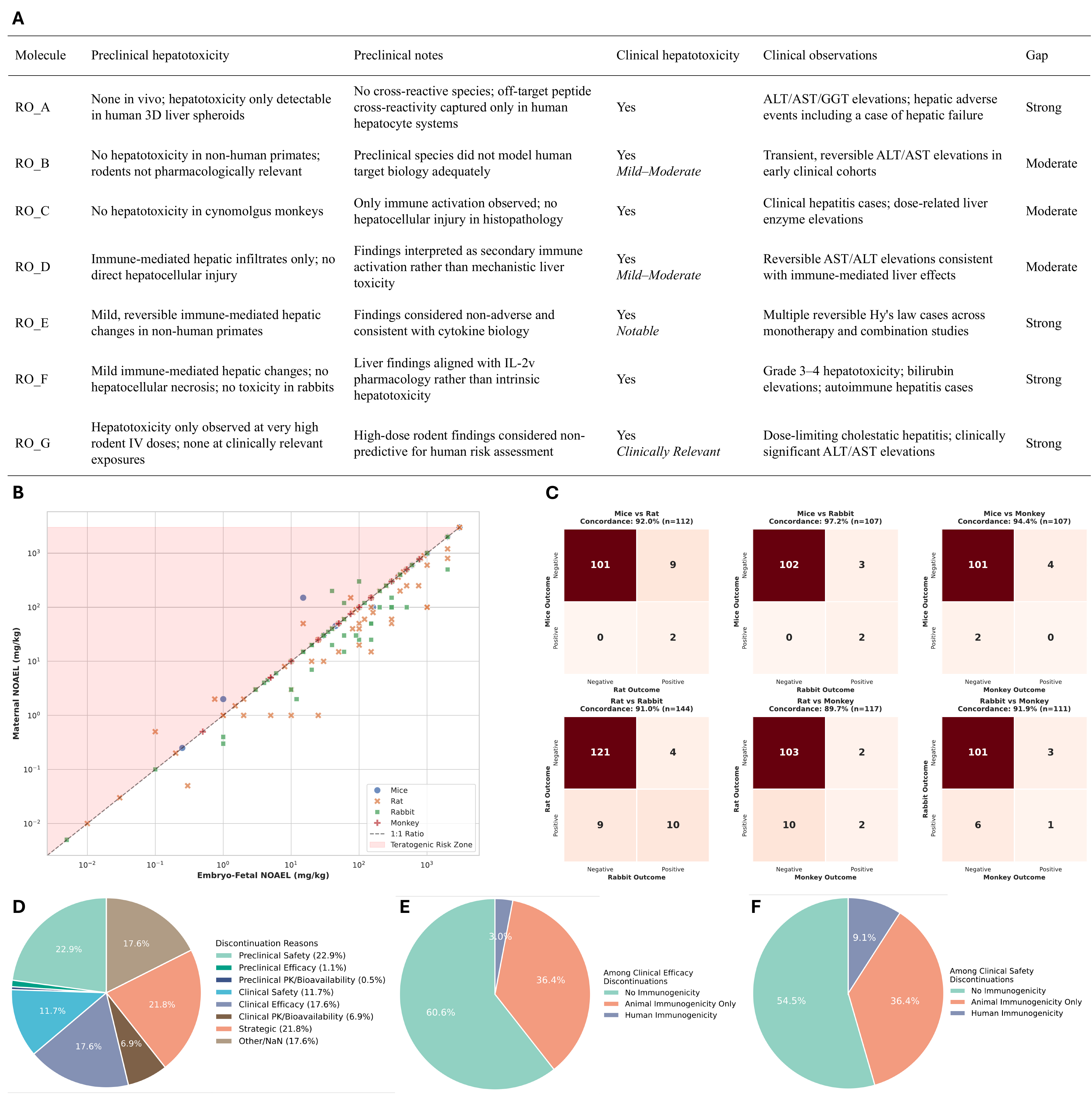}
    \caption{The Results of Reverse Translation, Quantitative Safety Assessment, and Immunogenicity Insights Enabled by \textit{DiscoVerse}; \textbf{A} Reverse Translation of Hepatotoxicity across Species; \textbf{B} Quantitative Risk Assessment for Maternal and Embryo-Fetal NOAEL; \textbf{C} Species Concordance Analysis for Teratogenicity; Retrospective Analysis of Immunogenicity in Discontinued Molecule Programs, \textbf{D} Discontinuation Reasons; \textbf{E} Clinical Efficacy by Immunogenicity; \textbf{F} Clinical Safety by Immunogenicity.}
    \label{fig:usecase}
\end{figure}

\paragraph{Use Case 1: Reverse Translation of Hepatotoxicity Across Species}

A recurring challenge in pharmaceutical research and development is understanding how safety signals observed in humans map back onto preclinical models, particularly for complex toxicities such as hepatotoxicity. Across therapeutic modalities including T-cell engagers, immune cytokine-based therapies, antisense molecules, and small molecules, preclinical programs sometimes show either minimal findings or rely on species that lack pharmacological relevance. As a result, clinically observed hepatotoxicity might emerge as a late finding, sometimes surprising development teams and complicating dose escalation, risk assessment, and patient monitoring.

\textit{DiscoVerse} directly addresses this challenge by enabling systematic reverse translation: given a clinical liver finding (e.g., LFT elevations, Hy’s law episodes, hepatocellular injury), the system retrieves, aligns, and synthesizes all available preclinical and in vitro evidence, including mechanistic insights, species limitations, biomarker signatures, and exposure-response relationships. This allows teams to determine whether the clinical observation was predictable, unexpected, or mechanistically explainable in hindsight. Applying \textit{DiscoVerse} across a set of discontinued and deprioritized molecules revealed a number of programs that showed clinical hepatoxicity findings and how those were captured preclinically.

Across these case studies, as shown in Fig.~\ref{fig:usecase}\textbf{A}, several cross-cutting scientific insights emerge. A major contributor to missed hepatotoxicity signals is \textbf{species non-relevance,} particularly for LM modalities (e.g., RO\_A and RO\_B), where no available animal species could model either the on-target biology or the off-target peptide--MHC interactions. \textit{DiscoVerse} makes such gaps immediately visible, prompting earlier reliance on human in vitro systems when LM modalities lack pharmacologically relevant in vivo models. A second pattern relates to \textbf{immune-mediated liver effects}, which were often mild or inconsistent in non-human primates yet emerged clinically for molecules such as RO\_C, RO\_E, and RO\_F. By integrating immune-activation signatures across heterogeneous documents, \textit{DiscoVerse }reveals these mechanistic connections more clearly than isolated study reports. A third insight concerns cases where \textbf{rodent findings were misleading predictors of human risk}: for example, RO\_G showed hepatotoxicity only at supratherapeutic exposures in rodents, obscuring its clinical liability. \textit{DiscoVerse} helps distinguish such high-dose toxicology artefacts from genuine translational signals. Finally, the analysis \textbf{underscores the value of human in vitro systems as early indicators of clinically relevant hepatotoxicity}. RO\_A exemplifies this, where 3D hepatocyte spheroid assays identified a clear hazard that no in vivo species could detect. By surfacing these patterns consistently, \textit{DiscoVerse} strengthens reverse translation and supports more informed, mechanism-aware safety assessment across programs.

This use case demonstrates how \textit{DiscoVerse} strengthens decision-making within research and development teams by uncovering hidden mechanistic patterns that link historical clinical events to preclinical biology, thereby accelerating risk assessment for new molecules targeting similar pathways. By systematically highlighting species gaps that warrant additional \textit{in vitro} or mechanistic investigations, \textit{DiscoVerse} ensures that non-relevance issues are recognised early rather than discovered retrospectively in clinical trials. At the same time, it provides structured and auditable evidence that integrates preclinical and clinical data accumulated over decades, transforming fragmented archival information into actionable insight. Beyond supporting individual project teams, \textit{DiscoVerse} also contributes to more effective pipeline governance by enabling earlier identification of modality-specific liabilities, clarifying translational risks, and informing strategic prioritisation or deprioritisation decisions. These insights help reduce downstream clinical safety failures, focus resources on assets with stronger translational confidence, and reinforce a proactive approach to risk mitigation across the portfolio. Collectively, these capabilities elevate reverse translation from an anecdotal, expert-memory--driven exercise to a systematic, evidence-based process, enabling more informed, consistent, and defensible decisions across discovery, early development, and portfolio strategy.

\paragraph{Use Case 2: Quantitative Risk Assessment in Embryo-Fetal Development} 

One persistent challenge in safety pharmacology is differentiating developmental toxicity secondary to maternal distress from \textbf{direct teratogenic effects driven by the molecule itself} at clinically relevant exposures. Making this distinction requires a consistent, quantitative comparison of maternal and embryo-fetal No-Observed-Adverse-Effect Levels (NOAELs), yet the necessary data are dispersed across a broad portfolio in inconsistent and unstructured formats. To establish whether a compound exerts embryo-specific toxicity independent of maternal compromise, we focused on calculating a Relative Sensitivity Ratio:
\begin{equation}
    \text{Relative Sensitivity Ratio} = \frac{\text{Maternal NOAEL}}{\text{Embryo-Fetal NOAEL}}
\end{equation}
Across an entire portfolio, this ratio can reveal mechanistic patterns that remain hidden when examining compounds one at a time. However, generating these safety margins at scale has traditionally been \textbf{computationally infeasible}, as the required NOAELs and associated pharmacokinetic parameters are embedded within unstructured narrative text. To overcome this barrier, we employed \textit{DiscoVerse} to perform systematic, quantitative data extraction across 180 molecules. The system decomposed complex toxicology queries to capture paired maternal and embryo-fetal NOAELs together with $C_{max}$ and AUC values across multiple species, including mice, rats, rabbits, and cynomolgus monkeys. Importantly, the Preclinical Agents went beyond simple numeric extraction: they triangulated evidence from discussion sections and results tables to validate specific malformations, filtering out non-specific toxicity and ensuring high-integrity inputs for downstream modeling. This structured extraction enabled computation of a robust Relative Sensitivity Ratio for each compound, revealing distinct safety clusters that had been obscured within static archival documents. As illustrated in Fig.~\ref{fig:usecase}\textbf{B}, the analysis highlighted a “Red Zone” (Relative Sensitivity Ratio $> 1$), where embryo-fetal toxicity occurred at exposures below the maternal NOAEL. In this case, it strongly indicates direct teratogenic liability rather than secondary physiological stress. By surfacing these quantitative patterns, \textit{DiscoVerse} converts previously inaccessible archival data into a predictive asset, enabling safety teams to pinpoint legacy compounds with embryo-specific risks and apply these mechanistic insights to the evaluation of emerging molecular scaffolds.

Having quantified molecule-driven teratogenic liability within species, a second challenge arises: evaluating how consistently these toxicological outcomes translate across species, a prerequisite for validating animal models used in developmental toxicity assessment. As shown in Fig.~\ref{fig:usecase}\textbf{C}, understanding the frequency with which compounds manifest teratogenic, embryolethal, or broader developmental toxicity findings in one species versus another is essential for determining model predictivity. To facilitate this, \textit{DiscoVerse} extracted binary outcomes (Positive/Negative) for mice, rats, rabbits, and cynomolgus monkeys from legacy reports, enabling direct calculation of species concordance rates (e.g., Rat vs. Rabbit) without manual cross-referencing. This harmonized dataset allows safety scientists to quantify alignment between models, identify species-specific sensitivities, and make more informed decisions about translational relevance early in development.

\paragraph{Use Case 3: Strategic Retrospective on Immunogenicity and Attrition}

Immunogenicity can fundamentally influence safety, efficacy, and molecule design strategies. However, because the discontinued portfolio contains a mix of large molecules, small molecules, and other modalities, it is important to recognize that immunogenicity, particularly in preclinical species, is frequently expected and not typically a reason for discontinuation. Preclinical immunogenicity in large molecules is common and rarely predictive of human outcomes, and true clinical discontinuations driven directly by human immunogenicity are exceptionally rare. In many cases, immunogenicity is a coincidental finding rather than the causative driver of safety or efficacy failure. It is interesting, however, to know whether automated analysis of historical development data can reliably determine if immunogenicity played a causative role in Clinical Safety or Clinical Efficacy discontinuations, and how frequently human versus animal immunogenicity events occur within these subsets.

With this context, we used \textbf{DiscoVerse} to stratify discontinued programs by their primary attrition rationale. As illustrated in Fig.~\ref{fig:usecase}\textbf{D}, the system successfully segmented the portfolio into distinct clusters, identifying Clinical Safety ($11.7\%$) and Clinical Efficacy ($17.6\%$) as key subsets alongside Strategic ($21.8\%$) and Preclinical Safety ($22.9\%$) drivers. To further investigate potential anti-drug antibody signals, we performed a targeted sub-analysis on the cohorts discontinued due to Clinical Efficacy (Fig.~\ref{fig:usecase}\textbf{E}) and Clinical Safety (Fig.~\ref{fig:usecase}\textbf{F}). DiscoVerse agents retrieved and synthesized clinical study reports to differentiate between animal and human immunogenicity events within these specific subsets. As shown in the Fig.~\ref{fig:usecase}\textbf{E} and~\ref{fig:usecase}\textbf{F}, while animal immunogenicity was frequently observed ($36.4\%$ in both cohorts), confirmed human immunogenicity was rare. The system identified molecules in the safety cohort and in the efficacy cohort with positive human immunogenicity. Beyond quantitative stratification, \textit{DiscoVerse} pinpointed the specific legacy molecules and studies associated with these signals, providing the Immunogenicity team with immediate access to the relevant context, which is information that would otherwise remain obscured due to the prohibitive time cost of manually screening historical archives.

\section{Discussion and Conclusion}
This work demonstrates that \textit{DiscoVerse} can transform archival pharmaceutical documentation into traceable, actionable knowledge. Traditional knowledge-management tools in research and development have struggled with data fragmentation, inconsistent terminologies, and loss of institutional memory. By decomposing complex scientific questions into domain-aligned sub-tasks (preclinical, clinical, strategic) and enforcing source-grounded synthesis through an orchestrating supervisor agent, \textit{DiscoVerse} operationalizes human-like division of labor within an automated, explainable framework. This mirrors the real-world workflow of interdisciplinary scientific teams, where subject-matter experts contribute specialized reasoning before results are merged into a coherent interpretation.

Across seven quantitative benchmarks (Q1–Q7), \textit{DiscoVerse} achieved near-perfect recall ($\geq 0.986$) with moderate precision (0.71–0.91) which is a favorable balance in safety-critical settings where false negatives are costlier than reviewing surplus candidates. False positives mainly stemmed from contextual ambiguities (e.g., confusing preclinical with clinical data) rather than hallucination, illustrating that retrieval breadth exceeded precision but remained evidence-linked and reviewable. Qualitative evaluations on Q8 (Discontinuation Rationale) and Q9 (Multi-Phase Toxicity Evidence Integration) further underscore the framework’s strength in multi-phase evidence integration, synthesizing disparate, evolving findings into structured, auditable summaries such as cross-species toxicity matrices. The examples shown demonstrate how \textit{DiscoVerse} disentangles overlapping data streams to produce scientifically faithful narratives while explicitly documenting provenance.

A defining feature of \textit{DiscoVerse} is its alignment with expert oversight. Instead of treating the system as an autonomous decision-maker, it is positioned as an augmentation layer, which automates exhaustive search and first-pass synthesis so human scientists can focus on contextual adjudication. This design mitigates hallucination risks and embeds regulatory traceability, key for pharmaceutical contexts subject to GLP/GVP compliance. Through source-linked outputs and structured schemas, \textit{DiscoVerse} makes “negative” or discontinued programs analyzable, turning decades of latent data into reusable assets for reverse translation. By bridging preclinical and clinical evidence, it enables hypothesis refinement, target validation, and safety-margin estimation based on historical outcomes, which is an essential step toward cyclical, data-driven learning across the drug development lifecycle.

Current limitations include (i) reliance on VLM-based parsing algorithm's quality for legacy scanned documents; (ii) moderate precision due to context drift and incomplete metadata; and (iii) dependence on manual adjudication for final interpretation. Future work will focus on context-aware rerankers, agentic feedback loops trained from expert annotations, and causal-chain reasoning modules that capture mechanistic relationships (e.g., dose-response-toxicity pathways). Expanding the evaluation to active development programs and integrating \textit{DiscoVerse} outputs with in-silico modeling pipelines could further validate its translational impact. Finally, establishing standardized audit trails and uncertainty quantification will facilitate deployment in regulated environments.

In conclusion, \textit{DiscoVerse} exemplifies how multi-agent LLM systems can serve as co-scientists in high-stakes, evidence-rich domains. By prioritizing grounding, explainability, and expert supervision over raw automation, it offers a blueprint for responsible AI integration into pharmaceutical research and development. When combined with institutional expertise, \textit{DiscoVerse} converts historical archives into living knowledge, which accelerates discovery, enhances reproducibility, and ensures that the scientific lessons of the past continually inform the medicines of the future.

\clearpage
\newpage
\bibliographystyle{unsrt}
\bibliography{ref}

%%%%%%%%%%%%%%%%%%%%%%%%%%%%%%%%%%%%%%%%%%%%%%%%%%%%%%%%%%%%
\clearpage
\newpage
\appendix

\section{Document Processing Pipeline and Retriever}
\label{rag}

\begin{figure}[h]
    \centering
    \includegraphics[width=0.8\linewidth]{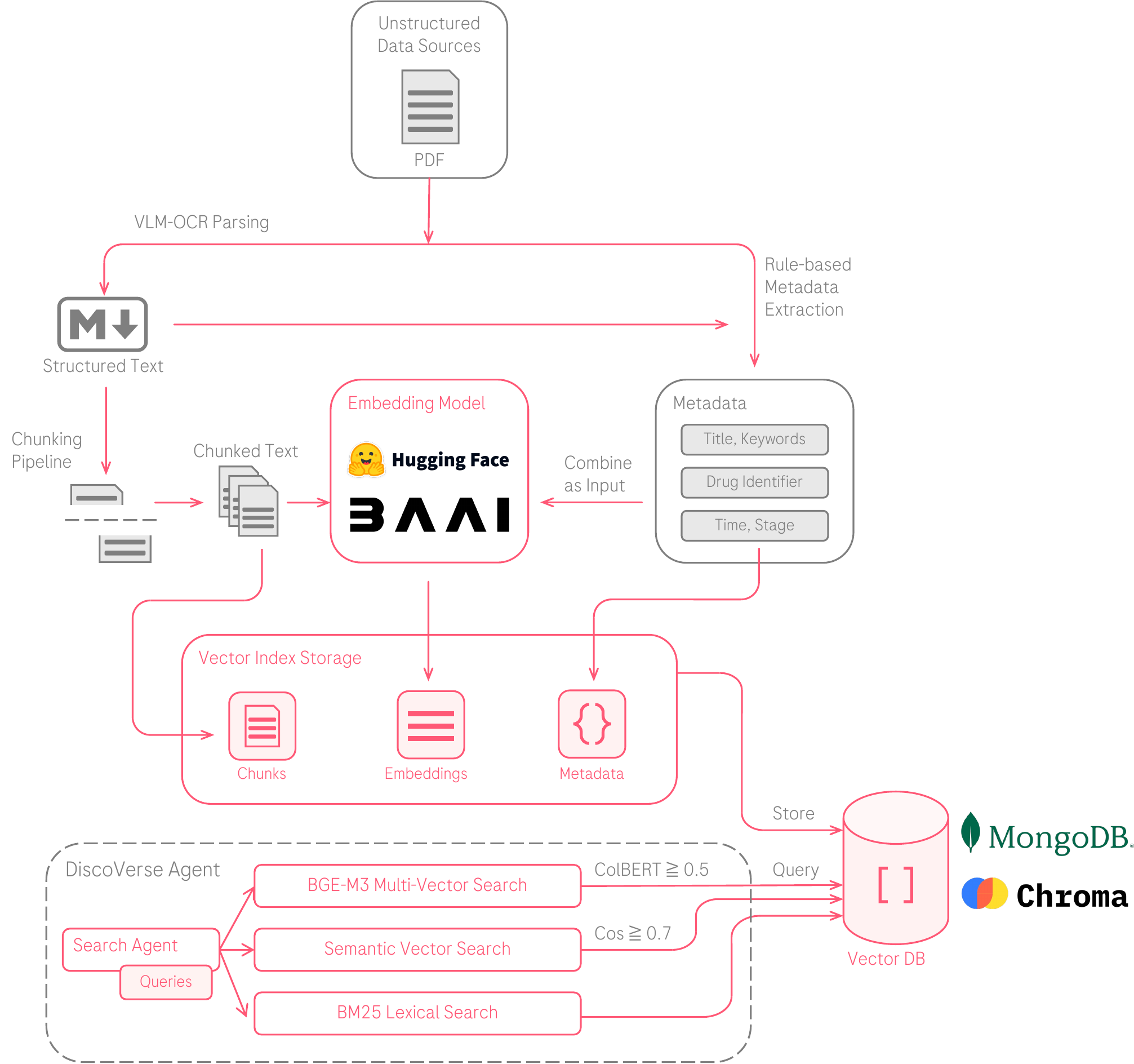}
    \caption{The illustrative overview of document processing and retriever in \textit{DiscoVerse}.}
    \label{fig:rag}
\end{figure}

\subsection{PDF Parsing}
We employed a Vision-Language Model (VLM)-based Optical Character Recognition (OCR) system, specifically olmOCR, for parsing PDF documents. This approach preserves the hierarchical structure inherent in any kind of papers, including section and subsection organization, text formatting attributes (bold, italic, etc.), mathematical formulas and equations, tables and structured data, citation references.

\subsection{Metadata Extraction}
Following the parsing stage, we extract metadata from two sources: (1) the parsed markdown content, and (2) the raw database records. Extracted metadata includes study title, drug identifier, temporal information, study stage, and relevant keywords. This metadata is then embedded alongside the corresponding text chunks during the embedding process, enriching the semantic representation of each chunk with structured contextual information.

\subsection{Text Chunking}
Documents were segmented into chunks of 512 words with a 64-word overlap between consecutive chunks. The overlap preserves contextual continuity across chunk boundaries.

To maintain semantic coherence, our chunking implementation respects section boundaries. Chunks do not break in the middle of logical sections. This section-aware approach ensures that retrieved passages maintain their original contextual structure.

\subsubsection{Embedding Models}
We utilized two multilingual embedding models to capture semantic representations of text chunks:
\begin{enumerate}
    \item intfloat/multilingual-e5-large-instruct: A multilingual embedding model with instruction-following capabilities, enabling task-specific encoding through natural language instructions.

    \item BGE-M3: A hybrid embedding model supporting dense retrieval, lexical matching, and multi-vector (ColBERT-style) representations within a unified framework.
\end{enumerate}

\subsection{Hybrid Retrieval System}
We implemented a hybrid retrieval system combining three retrieval mechanisms:

\begin{enumerate}
    \item Semantic Vector Search: Dense retrieval using the multilingual-e5-large-instruct embeddings with a similarity threshold of 0.7
    \item BGE-M3 Multi-Vector Search: ColBERT-style late interaction retrieval with a minimum ColBERT score threshold of 0.5
    \item BM25 Lexical Search: Keyword-based retrieval for exact term matching
\end{enumerate}

Results from all three retrievers are merged and deduplicated to produce the final candidate set. This hybrid approach combines semantic understanding with lexical precision.

The retrieval infrastructure uses two storage backends: ChromaDB (an open-source vector database) for vector similarity search operations and MongoDB (deployed on-premises) for document storage and metadata management.

%%%%%%%%%%%%%%%%%%%%%%%%%%%%%%%%%%%%%%%%%%%%%%%%%%%%%%%%%%%%
\section{Evaluation Design for Queries}
\label{expert_eva}
Each LLM output was manually reviewed against its source document and classified into one of four categories: True Positive (TP), True Negative (TN), False Positive (FP), or False Negative (FN). The evaluation criteria were tailored to each of the seven queries for all 180 drugs as detailed below.

\subsection{Query 1: First-in-Human (FIH) Dose}

\textbf{Query:} "What was the first in human dose for (molecule)?"
\paragraph{Positive Case:} The document explicitly states the FIH dose.
\paragraph{Negative Case:} The document discusses the molecule but does not mention the FIH dose.
\paragraph{Classification Criteria:}
\begin{itemize}
    \item \textbf{True Positive (TP):} The LLM correctly extracts the FIH dose and its accompanying context is factually correct. For example, extracting "The FIH was 5 mg/kg". If context is given (e.g. the correct study population and trial phase) it also has to be correct.
    \item \textbf{False Positive (FP):} The LLM extracts an incorrect dose, such as a preclinical dose from an animal study, or hallucinates a value not present in the text.
    \item \textbf{False Negative (FN):} The LLM fails to find the FIH dose even though it is stated in the document.
    \item \textbf{True Negative (TN):} The LLM correctly reports that the information is not available in the provided text.
\end{itemize}

\subsection{Query 2: Route of Administration (RoA)}

\textbf{Query:} "What was the route of administration in humans for drug X?"
\paragraph{Positive Case:} The document mentions how the drug was administered to humans (e.g., oral, intravenous).
\paragraph{Negative Case:} The route of administration is not mentioned.
\paragraph{Classification Criteria:}
\begin{itemize}
    \item \textbf{True Positive (TP):} The LLM correctly identifies the route of administration.
    \item \textbf{False Positive (FP):} The LLM states the wrong route or confuses a preclinical (animal) route with the human route.
    \item \textbf{False Negative (FN):} The LLM fails to find the RoA even though it is stated.
    \item \textbf{True Negative (TN):} The LLM correctly reports that the RoA is not mentioned.
\end{itemize}
 
\subsection{Query 3: Highest Dose in Phase I(MAD)/II}

\textbf{Query:} "What was the highest clinical dose in Phase I(MAD)/II for (molecule)?"
\paragraph{Positive Case:} The document contains dose information for Phase I Multiple Ascending Dose (MAD) and/or Phase II studies.
\paragraph{Negative Case:} The document does not contain this specific dose information.
\paragraph{Classification Criteria:}
\begin{itemize}
    \item \textbf{True Positive (TP):} The LLM correctly identifies the single highest dose administered across all relevant study phases mentioned.
    \item \textbf{False Positive (FP):} The LLM extracts a dose from an incorrect phase (e.g., Phase III), fails to identify the maximum value among several options, or incorrectly identifies a dose as being from a MAD study.
    \item \textbf{False Negative (FN):} The LLM reports that the information is not found despite it being present.
    \item \textbf{True Negative (TN):} The LLM correctly determines that no dose information for Phase I(MAD) or Phase II is available.
\end{itemize}

\subsection{Query 4: Highest Dose with Severe Adverse Events (SAEs)}

\textbf{Query:} "What was the highest clinical dose at which there were severe adverse events for drug (molecule)?"
\paragraph{Positive Case:} The document explicitly links a dose level to the occurrence of SAEs, Dose-Limiting Toxicities (DLTs), or Grade 3+ adverse events.
\paragraph{Negative Case:} The document reports no SAEs or does not link them to a specific dose.
\paragraph{Classification Criteria:}
\begin{itemize}
    \item \textbf{True Positive (TP):} The LLM correctly extracts the dose that is explicitly associated with an SAE or DLT.
    \item \textbf{False Positive (FP):} The LLM incorrectly associates a dose with an unrelated SAE or confuses a mild adverse event with a severe one. This represents a critical failure in relational extraction.
    \item \textbf{False Negative (FN):} The LLM fails to connect a dose and an SAE even when the relationship is clearly stated.
    \item \textbf{True Negative (TN):} The LLM correctly reports that no dose was explicitly linked to SAEs.
\end{itemize}

\subsection{Query 5: Efficacious Dose}

\textbf{Query:} "What was the efficacious dose in the clinic?"
\paragraph{Positive Case:} The document discusses clinical efficacy, pharmacodynamic markers, or therapeutic response at specific doses.
\paragraph{Negative Case:} The document only discusses safety and/or pharmacokinetics, not efficacy.
\paragraph{Classification Criteria:}
\begin{itemize}
    \item \textbf{True Positive (TP):} The LLM correctly extracts the efficacious dose(s) and accurately summarizes the associated outcomes and patient populations.
    \item \textbf{False Positive (FP):} The LLM extracts a dose but misrepresents its efficacy (e.g., claims a dose was effective when the trial failed to show a benefit).
    \item \textbf{False Negative (FN):} The LLM fails to find the efficacy information despite it being present in the text.
    \item \textbf{True Negative (TN):} The LLM correctly reports that no efficacy data is available.
\end{itemize}

\subsection{Query 6: Treatment Regimen}

\textbf{Query:} "What was the treatment regimen for drug X in humans?"
\paragraph{Positive Case:} The document describes the complete dosing schedule (dose level, frequency, duration).
\paragraph{Negative Case:} The regimen details are not described.
\paragraph{Classification Criteria:}
\begin{itemize}
    \item \textbf{True Positive (TP):} The LLM correctly synthesizes the dose level(s), frequency, duration, and patient population into a coherent summary.
    \item \textbf{False Positive (FP):} The LLM gets one or more elements of the regimen wrong (e.g., mistakes a single-dose study for a multiple-dose study).
    \item \textbf{False Negative (FN):} The LLM fails to assemble the regimen details from the text.
    \item \textbf{True Negative (TN):} The LLM correctly reports that the regimen is not detailed.
\end{itemize}

\subsection{Query 7: Margin of Safety}

\textbf{Query:} "What do we know about the Margin of Safety of drug X in the first IB?"
\paragraph{Positive Case:} The document contains preclinical toxicology data (e.g., NOAEL) and the clinical starting dose.
\paragraph{Negative Case:} This information is not present to calculate or assess the safety margin.
\paragraph{Classification Criteria:}
\begin{itemize}
    \item \textbf{True Positive (TP):} The LLM correctly synthesizes preclinical safety data and clinical dose data to accurately describe the safety margin.
    \item \textbf{False Positive (FP):} The LLM uses incorrect numbers (e.g., a toxic dose instead of a no-effect dose), leading to a miscalculation or misrepresentation of the safety margin.
    \item \textbf{False Negative (FN):} The necessary data is present, but the LLM fails to connect them to describe the safety margin.
    \item \textbf{True Negative (TN):} The LLM correctly reports that the data needed to assess the safety margin is absent.
\end{itemize}
%%%%%%%%%%%%%%%%%%%%%%%%%%%%%%%%%%%%%%%%%%%%%%%%%%%%%%%%%%%%
\section{Evaluation Metrics}
\label{metrics}
The following metrics were calculated for each of the seven benchmark queries:

\textbf{Accuracy}: The overall proportion of correct predictions, calculated as

\begin{equation}
    \text{Accuracy} = \frac{TP + TN}{TP + TN + FP + FN}
\end{equation}

\textbf{Precision (Positive Predictive Value)}: The proportion of positive predictions that were factually correct, calculated as  

\begin{equation}
    \text{Precision} = \frac{TP}{TP + FP}
\end{equation}

This metric is a direct measure of the system's factuality and its tendency to avoid hallucinations or contextually incorrect extractions.

\textbf{Recall (Sensitivity)}: The proportion of all actual positive cases that were correctly identified by the system, calculated as 

\begin{equation}
    \text{Recall} = \frac{TP}{TP + FN}
\end{equation}

This metric measures the agent's completeness and its ability to successfully find the "needle" when it is present.

\textbf{Specificity}: The proportion of all actual negative cases that were correctly identified, calculated as

\begin{equation}
    \text{Specificity} = \frac{TN}{TN + FP}
\end{equation}

This measures the agent's ability to correctly recognize the absence of information.

\textbf{F1-Score}: The harmonic mean of Precision and Recall, calculated as 

\begin{equation}
    \text{F1-Score} = 2\times\frac{\text{Precision} \times \text{Recall}}{\text{Precision} + \text{Recall}}
\end{equation}

It provides a single, balanced score that accounts for both factuality and completeness.

\end{document}